\documentclass[10pt,twocolumn,letterpaper]{article}

\usepackage{cvpr}
\usepackage{times}
\usepackage{epsfig}
\usepackage{graphicx}
\usepackage{amsmath}
\usepackage{amssymb}
\usepackage{booktabs}
\usepackage{multirow}

\usepackage[pagebackref=true,breaklinks=true,letterpaper=true,colorlinks,bookmarks=false]{hyperref}

 \cvprfinalcopy 

\def\cvprPaperID{7942} 

\ifcvprfinal\fi
\begin{document}

%
\title{Joint Semantic Segmentation and Boundary Detection using\\ Iterative Pyramid Contexts }
\author{Mingmin Zhen$^{1}$\hspace{1.0cm} Jinglu Wang$^{2}$\hspace{0.7cm} Lei Zhou$^{1}$\hspace{0.7cm} Shiwei Li$^{3}$ \\ Tianwei Shen$^{1}$\hspace{0.7cm} Jiaxiang Shang$^{1}$\hspace{0.7cm} Tian Fang$^{3}$\hspace{0.7cm} Long Quan$^{1}$ \\
\normalsize $^1$Hong Kong University of Science and Technology\hspace{0.7cm}  \normalsize $^2$Microsoft Research Asia\hspace{0.7cm} \normalsize $^3$Everest Innovation Technology \\
	\tt\small\{mzhen, lzhouai, tshenaa, jshang, quan\}@cse.ust.hk \\
	\tt\small Jinglu.Wang@microsoft.com\hspace{0.7cm} \{sli, fangtian\}@altizure.com}


\maketitle

\begin{abstract}
In this paper, we present a joint multi-task learning framework for semantic segmentation and boundary detection. The critical component in the framework is the iterative pyramid context module (PCM), which couples two tasks and stores the shared latent semantics to interact between the two tasks. For semantic boundary detection, we propose the novel spatial gradient fusion to suppress non-semantic edges. As semantic boundary detection is the dual task of semantic segmentation, we introduce a loss function with boundary consistency constraint to improve the boundary pixel accuracy for semantic segmentation. Our extensive experiments demonstrate superior performance over state-of-the-art works, not only in semantic segmentation but also in semantic boundary detection.  In particular, a mean IoU score of $81.8\%$  on Cityscapes test set is achieved without using coarse data or any external data for semantic segmentation. For semantic boundary detection, we improve over previous state-of-the-art works by  $9.9\%$ in terms of AP  and $6.8\%$ in terms of MF(ODS). 
\end{abstract}

\section{Introduction}
\begin{figure}[t]
	\centering
	\includegraphics[width=\linewidth]{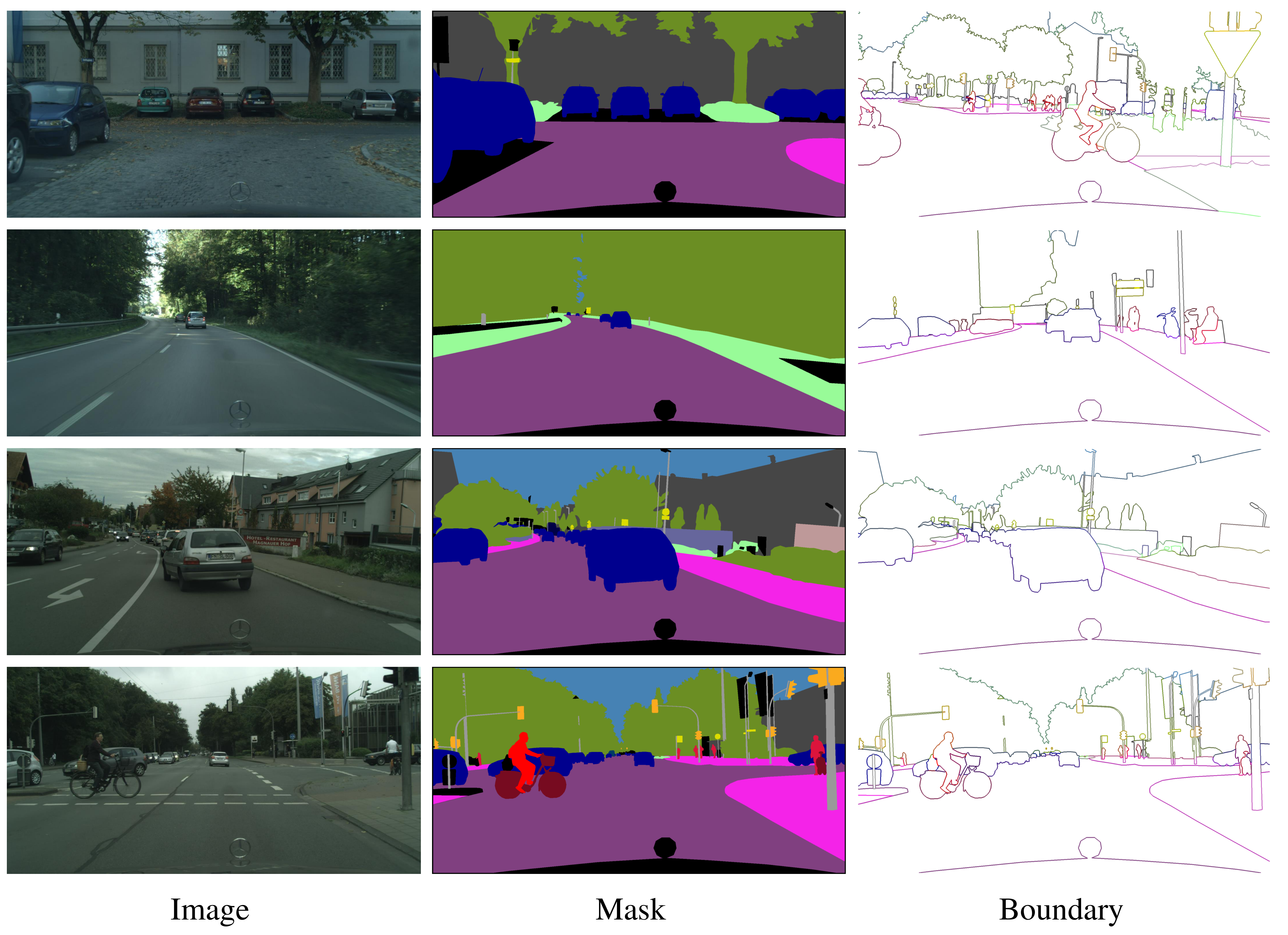}
	\caption{Some samples  for  semantic segmentation in Cityscape dataset (\textbf{best viewed in color}). We follow \cite{casenet,steal} to generate the semantic boundary.}
	\label{cs_mask_edge}
\end{figure}

Semantic segmentation has been actively studied in many recent papers and is also critical for various challenging applications such as autonomous driving \cite{auto_drive} and virtual reality \cite{vr}. In semantic segmentation tasks, we  estimate a mask where each pixel represents a category ID (Figure \ref{cs_mask_edge}).  The semantic boundary detection task  is a multi-label  classification task and different from traditional binary edge detection. As a dual problem of semantic segmentation, which means that the boundary always surrounds the mask, the goal of  semantic boundary detection \cite{casenet,steal} is to  identify image pixels that belong to object (class) boundaries.  In general, estimating the semantic label at image boundaries is challenging as it could  be ambiguous between two sides. The boundary accuracy of this mask is crucial to the final semantic segmentation accuracy, yet its importance is often overlooked in previous methods.   \\
\indent Recently, some works  adopt edge detetion as auxiliary information to improve the performance of semantic segmentation. In \cite{gate_scnn}, a two-stream CNN architecture for semantic segmentation is proposed that explicitly wires shape information as a separate processing branch, i.e., shape stream, that processes information in parallel to the classical stream. \cite{BFP} learns the edges as an additional semantic class to enable the network to be aware of the boundary layout. Unidirectional acyclic graphs (UAGs) are used to model the function of undirected cyclic graphs (UCGs) in order to overcome the loopy property and improve efficiency. Based on the proposed UAGs, holistic context is aggregated via harvesting and propagating the local features throughout the whole image efficiently. However, these works ignore the semantic information of edges in semantic segmentation. In essence, semantic boundary detection is more highly coupled with semantic segmentation compared with  binary edge detection.  \\
\indent  In this paper, we take a step further to propose a joint-task framework combining both semantic segmentation and semantic boundary detection. 
Semantic segmentation task and semantic boundary task are correlated together iteratively, which conforms to the dual relationship between them. We leverage pyramid context information of one task to refine another task. \\
\indent For semantic boundary detection, one challenging issue is to suppress the non-semantic edges, which are ambiguous to distinguish from semantic edges. To address this problem, we derive the spatial gradient, which is similar to the image gradient method \cite{canny_edge},  as the initial semantic boundary from the semantic mask and fuse it with the semantic boundary probability map to obtain clean semantic boundary results.  \\
\indent As there exists the duality constraint between semantic segmentation and semantic boundary detection,  we propose a novel loss function to enforce boundary consistency for semantic segmentation task. For the predicted mask, boundary is derived as the outer contour, which can be used to constrain the mask. The differences between the prediction results and the groundtruth are then formulated as one loss term, which we refer to as duality loss,  to impose boundary consistency on semantic mask during model training. The duality loss term is differentiable due to the pixel-wise operation. The overall network can be trained in an end-to-end manner. \\
\indent Our experiments show that our proposed method outperforms existing state-of-the-art works, especially for semantic boundary detection task. The proposed method surpasses  state-of-the-art work \cite{steal} by 9.9\% in terms of average precision (AP) and 6.8\% in terms of maximum F-measure (MF) at optimal dataset scale (ODS) for semantic boundary detection.  We achieve mean intersection over union (mIoU)   score $81.8\%$ on Cityscape test set with only fine annotated trainval data used for training.   The performance on the validation set is also better than previous works. The significant gains of performance verify the effect of proposed method. All in all, our main contributions can be summarized as follows:
\begin{itemize}
	\item To our best knowledge, we are the \textbf{first} to  combine semantic boundary detection task and semantic segmentation task    into a joint multiple-task learning framework with iterative pyramid context module (PCM).   
	\item In the semantic boundary detection module,  we introduce a novel strategy to suppress non-semantic edges, by fusing  the derived boundary from mask probability map with semantic boundary probability map.
	\item In the semantic segmentation module, We design a duality loss, which improves the boundary pixel accuracy by enforcing consistency between  boundary derived from the mask and boundary groundtruth.
\end{itemize}
\begin{figure*}[t!]
	\centering
	\includegraphics[width=\linewidth]{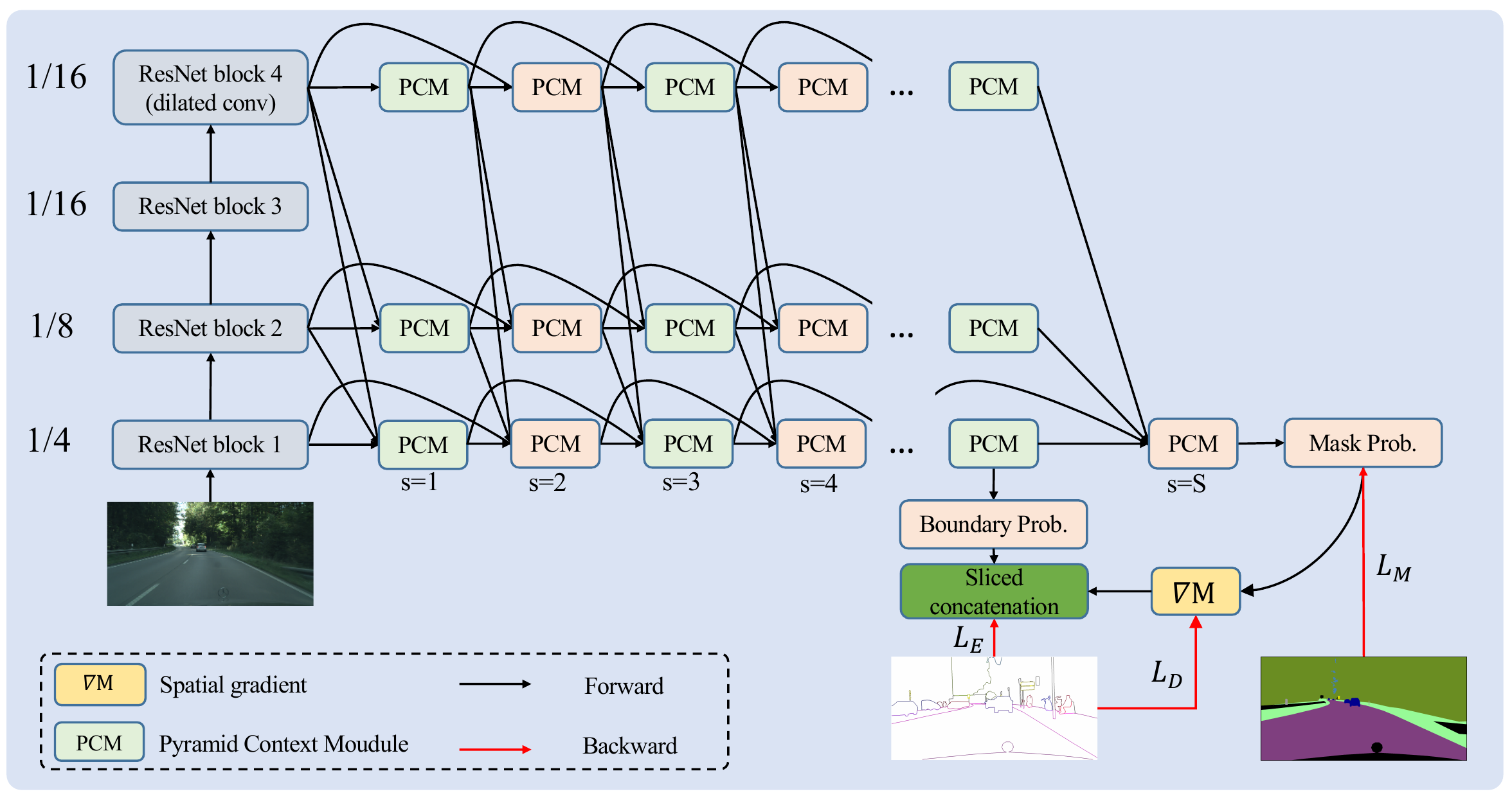}
	\caption{Overview of  RPCNet  for semantic segmentation and semantic boundary detection. The iterative pyramid context module (PCM) is used to interact between the two tasks. We leverage the context from one task to refine the feature map from  another task alternately.  The spatial gradient  $\nabla M$ derived from semantic mask is fused together with the probability map from semantic boundary  through sliced concatenation. The duality loss is also applied on semantic segmentation to improve the boundary accuracy of semantic mask. 
	}
	\label{pcnet}
\end{figure*}
\section{Related work}
\noindent \textbf{Semantic segmentation} Fully convolutional network (FCN) \cite{fcn_cvpr,fcn_tpami} based methods \cite{lfov,deconvnet,u-net,LRR,gateFrNet,sdn,parsenet,fdnet} make great progress in image semantic segmentation. In \cite{lfov}, last two downsample layers are removed to obtain dense prediction and dilated convolution operations are employed to enlarge the receptive field. DenseCRF \cite{densecrf,lfov} is also used to capture long range dependencies between pixels. After that, an end-to-end CRF based method CRF-RNN is proposed to refine the semantic segmentation result.  Unet \cite{u-net}, Deeplabv3+ \cite{deeplabv3+}, RefineNet \cite{refinenet} and DFN \cite{dfn} adopt encoder-decoder structures that fuse the information in low-level and high-level layers to predict segmentation mask. \\
\indent  Recently, some works attempt to improve the feature representation ability through aggregating the contextual information. In \cite{deeplabv3}, the ASPP module is used to capture contextual information by using different dilation convolutions. PSPNet \cite{PSPNet} introduces pyramid pooling over sub-regions of four pyramid scales, and all the pixels within the same sub-region are treated as the context for the pixels belonging to the sub-region. ParseNet \cite{parsenet} utilizes global pooling to harvest context information for global representations. Zhao et al. \cite{psanet} propose the pointwise spatial attention network which uses predicted attention map to guide contextual information collection. \\
\noindent \textbf{Semantic boundary detection}
Recently, \cite{casenet} extends the CNN based class-agnostic edge detector proposed in \cite{hed}, and allows each edge pixel to be associated with more than one class. The  CASENet \cite{casenet} architecture combines low and high-level features with a multi-label loss function to supervise the fused activations. Most works use non-maximum-suppression (NMS) \cite{canny_edge} as a postprocessing step in order to deal with the thickness of predicted boundaries. In \cite{steal}, a simple and effective Thinning Layer and loss that can be used in conjunction with existing boundary detectors is proposed.\\
\indent The most related works to this paper are Gated-SCNN \cite{gate_scnn} and BFP \cite{BFP}. In \cite{gate_scnn}, Towaki et al. adopt the two-stream architecture, including shape stream and classical mask stream, with gate design to facilitates the flow of information from the regular stream to the shape stream. Similarly, boundary is learned as an additional semantic class to enable the network to be aware of the boundary layout in \cite{BFP}. Additionally, boundary aware feature propagation (BFP) module is proposed to harvest and propagate the local features within their regions isolated by the learned boundaries in the UAG-structured image. In this paper, we employ the semantic boundary to help refine semantic segmentation results, which is much more consistent with semantic segmentation compared with binary boundary information. The two tasks are complementary to each other iteratively.   
In addition, we constrain semantic segmentation by using outer contour of semantic mask to compute duality loss, which improves   boundary accuracy of semantic mask.
\section{Approach}
In this section, we present the proposed iterative pyramid context network (RPCNet) in detail, which is illustrated in Figure \ref{pcnet}. We first give an overview of the whole architecture in section \ref{arthitecture}. Next, the iterative pyramid context module (PCM), which captures the pyramid context from multi-scale feature maps, is elaborated in section \ref{rpcm}. Then we introduce details about the spatial gradient fusion in section \ref{sg_fusion}. At last, we describe the computation process of  duality loss in the  framework in section \ref{du_loss}.
\subsection{Architecture}\label{arthitecture}
In this paper, we employ a pretrained residual network \cite{resnet_he}  with the dilated strategy \cite{deeplab_v2} as the backbone. Note that we remove the downsampling operations and employ dilation convolutions in the last  ResNet blocks, thus enlarging the size of the final feature map size to 1/16 of the input image. This retains more details without adding extra parameters. The feature maps with different scales from ResNet101 backbone are firstly fed into a $3 \times 3$ convolution followed by ReLU, batch normalization (BN) layer to reduce feature map number to 256, and the outputs are then taken as input of the iterative pyramid context module. We perform task-interaction by pyramid context refining on multiple levels. At each level,  the two tasks are performed alternately. We leverage high-level feature maps, including the same level feature map, to refine the low-level feature maps. \\
\indent More formally, let $s$ denote the step of pyramid context capturing ($1 \leq s \leq S$) as shown in Figure \ref{pcnet}, and $F_s^{t}$ denotes $s$-th step feature map at $t$-th level. We use three level feature maps from the backbone. Thus, $t$ is from $0$ to $2$. $F_s^{0}$, $F_s^{1}$ and $F_s^{2}$ stand for feature maps with $\frac{1}{16}$, $\frac{1}{8}$ and $\frac{1}{4}$ size. When $s=0$, $F_{0}^{t}$ denotes feature map extracted from the backbone. \\
\indent After total $S$ steps, we can obtain  fine feature maps for semantic segmentation and semantic boundary detection. For semantic boundary detection, we perform sliced concatenation operation on feature maps from semantic boundary detection task and feature maps containing semantic boundary information derived from semantic segmentation task. The spatial gradient is computed from the semantic segmentation task to obtain auxiliary semantic boundary probability map.  For semantic segmentation, we adopt the duality loss to improve the boundary accuracy of semantic mask. As semantic boundary is easily derived from the semantic mask, we compute the spatial derivative $\nabla M$ on the probability map to obtain initial semantic boundary. The initial semantic boundary is compared with groundtruth boundary to compute duality loss.
\begin{figure}[t]
	\centering
	\includegraphics[width=0.95\linewidth]{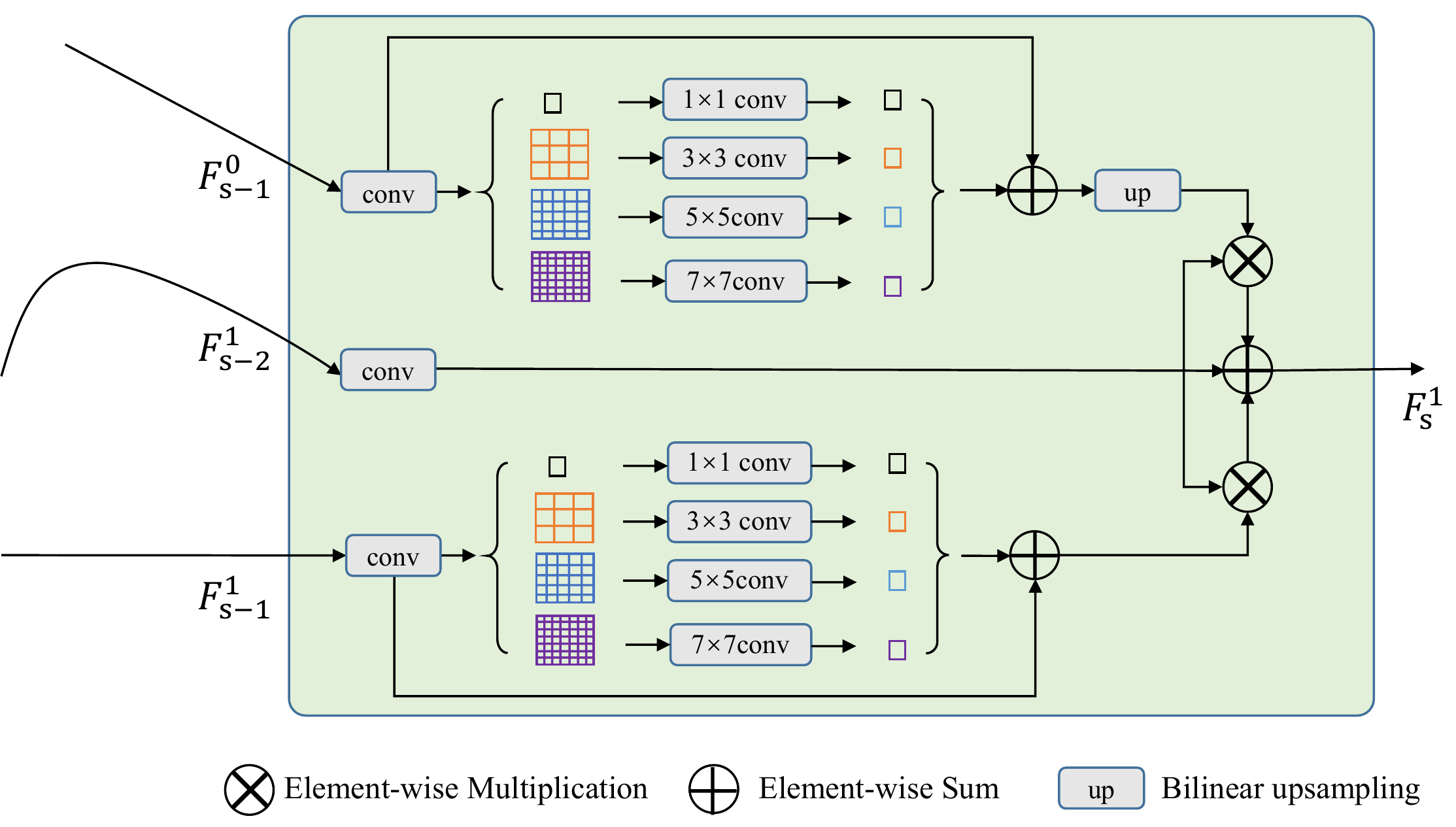}
	\caption{An example of pyramid context module (PCM) at  step $s=3$ and  level $t=1$. The feature maps from the higher level and the same level are firstly divided into multiple patches by global pooling. The global feature maps for different patches are then fed into a corresponding convolution layer to obtain the global context information. The feature map from one task (semantic segmentation or semantic boundary detection) is refined by these global context features. ``up'' operation uses bilinear interpolation method to upsample the input feature map into the same feature map size at level $t=1$. }
	\label{pcm}
\end{figure}
\subsection{Iterative Pyramid Context Module}\label{rpcm}
\indent We use the iterative pyramid context module to capture the global context from multiple levels to correlate the  two tasks, where global context from one task is to purify the feature maps from another task. \\
\indent For the input feature map $F_{s-2}^{t}$ at $s$-th step ($s \geq 2$) and $t$-level, we use higher level feature maps $F_{s-1}^{t'}$ ($t'$ is from $0$ to $t$) at $s-1$ step to refine it. When $s=1$, we refine $F_0^{t}$ by $F_0^{t'}$. We perform context capturing by dividing the feature map into $G \times G$ patches and computing global context for each patch. Specifically, given the feature map $F_{s-1}^{t'} \in \mathbb{R}^{H \times W \times C}$, we obtain patch based context $P_{G \times G}^{t'} \in \mathbb{R} ^{H \times W \times C}$ by 
 \begin{equation}
 	P_{G \times G}^{t'} (x, y) = \frac{1}{|S(x, y)|} \sum_{(h, w) \in S(x, y)}F(h, w)
 \end{equation}
 where $x=0,1,..., G-1$ and $y=0,1,..., G-1$. $S(x, y)$ is used to denote a patch located  at $x, y$. The  features for each patch are then connected together by a $G \times G$ convolution layer, which yields a $C$ dimensional feature vector $F_{G}^{t'}$ as shown in Figure \ref{pcm}. We sum up all these context features and input feature map $F_{s-1}^{t'}$ to get context feature map $F_{context}^{t'}$. After the context feature map $F_{context}^{t'}$ is upsampled to the same size with $F_{s-2}^{t}$, we perform element-wise multiplication  operation between them and sum up the refined feature map and input feature map $F_{s-2}^{t}$. Thus, we can obtain final output $F_{s}^{t}$ by 
 \begin{small}
 \begin{equation}
 \begin{split}
F_{s}^{t}  &=  F_{s-2}^{t} + \sum_{0 \le t' \le t} F_{s-2}^{t} \odot F_{context}^{t'} \\
&= F_{s-2}^{t} + \sum_{0 \le t' \le t} ( F_{s-2}^{t} \odot F_{s-1}^{t'} + \sum_{G \in \{1, 3, 5, 7\}} F_{s-2}^{t} \odot F_{G}^{t'})
 \end{split}
   \label{e_patch}
\end{equation}
\end{small}
where $\odot$ is element-wise multiplication and  we empirically use the 4 different patch setup  of $F_{s-1}^{t'}$ (i.e., $1 \times 1$, $3 \times 3$, $5 \times 5$ and $7 \times 7$ patches) to compute the set of feature maps $P_{G \times G}^{t'}$.  A special case is $t=0$,  where there is no higher-level feature map used. This feature map would simply undergo a $3 \times 3$ convolution with ReLU and BN layer.\\
\indent From Equation \ref{e_patch}, we can see that the input feature map $F_{s-2}^{t}$ is refined by pyramid context representations, which are from different level feature maps with different scales and different context collection from different patch partition. The context refinement helps to yield finer feature representation on the one hand. On the other hand, it boosts the interaction between semantic segmentation and semantic boundary detection. We also use  iterative pyramid context module to propagate contextual information between the two tasks, as shown in Figure \ref{pcnet}.  Thanks to the pyramid context module, the feature maps for the two tasks are closely correlated  and mutually collaborate to boost the performance.
\subsection{Spatial Gradient $\nabla M$ Fusion}\label{sg_fusion}
\begin{figure}[t]
	\centering
	\includegraphics[width=0.95\linewidth]{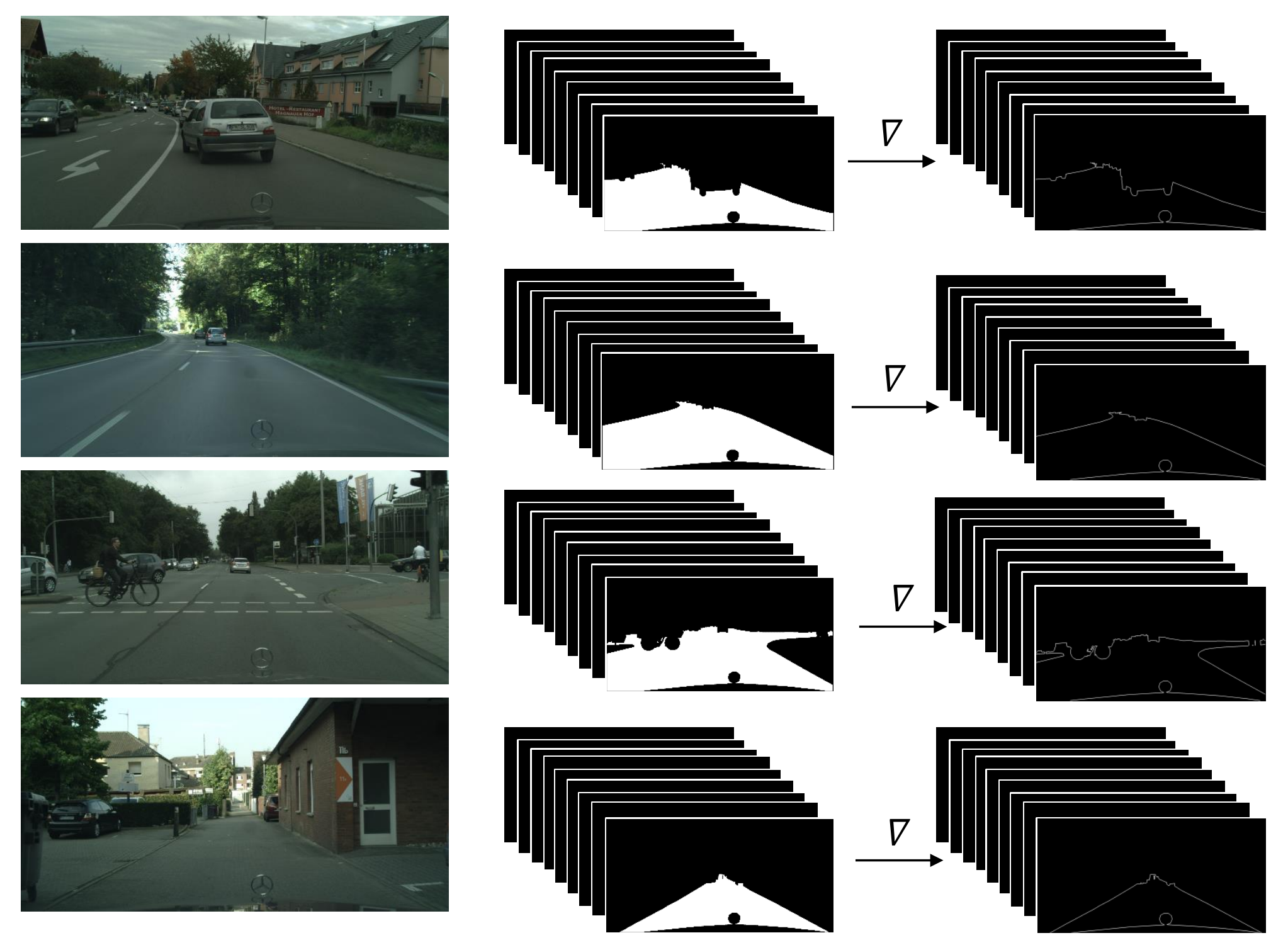}
	\caption{Some visualization examples for semantic mask and inferred semantic boundary through spatial gradient derivation on different categories (\textbf{best viewed in color}).}
	\label{mask2edge}
\end{figure}

After several steps of the pyramid context module, we can  obtain the semantic mask probability map $M \in \mathbb{R}^{H \times W \times K}$ and the semantic boundary probability map $B \in \mathbb{R}^{H \times W \times K}$, where $K$ is the number of categories. We can obtain semantic boundary from semantic segmentation mask easily by spatial gradient deriving. Here, we use adaptive pooling to derive spatial gradient $\nabla M$, which is
\begin{equation}
	\nabla M(x, y) = |M(x, y) - pool_k(M(x, y))|
\end{equation}
where $x$ and $y$ denote the location of mask probability map and $|. |$ remarks the absolute value function. $pool_k$ is an adaptive average pooling operation with kernel size $k$. $k$ is used to control the derived boundary width and is set to 3 in the framework. Some examples are shown in Figure \ref{mask2edge}. For the input semantic segmentation mask, we can obtain precise semantic boundary results. \\
 \indent To augment the semantic boundary detection, we fuse the boundary probability map $B = \{B_1, B_2, ..., B_K\}$ and inferred boundary map $\nabla M = \{\nabla M_1, \nabla M_2, ..., \nabla M_K\}$  into new boundary $B'$ with $2K$ channels by sliced concatenation operation:
\begin{equation}
	[B_1, \nabla M_1, B_2, \nabla M_2, ..., B_K, \nabla M_K].
\end{equation}
The resulting concatenated activation map is  fed into a convolution layer with K-grouped convolution to produce a K-channel probability map $Y \in \mathbb{R}^{H \times W \times K}$, which is then used to compute semantic boundary detection loss with corresponding groundtruth. \\
\indent  Semantic boundary detection suffers from coarse boundary in previous works \cite{casenet,steal} as the boundary is very sparse. In this paper, we fuse the inferred boundaries from the semantic mask to suppress non-edge pixels and localize detailed boundaries, which is desirable in semantic boundary detection.
\subsection{Duality Loss}\label{du_loss}
As shown in Figure \ref{pcnet}, there are two tasks for our proposed network. We compute two kinds of loss corresponding to the probability feature maps for the tasks. \\
\noindent \textbf{Semantic Mask  Loss} For semantic segmentation, it is common to  compute cross-entropy loss for each pixel, which can be referred to as $L_M$. As it treats all pixels equally, the pixels around the boundary, which are ambiguous, are inconsistent with the groundtruth. We introduce a duality loss for semantic segmentation, assuming that there exists a consistency between the boundaries of segmented object with the groundtruth of object boundary. \\
\indent  As we derive semantic boundary $\nabla M$ from semantic mask, we measure inconsistency between it and semantic boundary groundtruth. We compute $l_1$ loss for it as follows
\begin{equation}
	L_D = \sum_{i} |\nabla M_{i} - B_{i}^{gt}|
\end{equation}
where $B^{gt}$ is the semantic boundary groundtruth derived from semantic segmentation mask groundtruth. \\
\indent Two consistency constraints are imposed to enhance the performance of semantic segmentation. The cross-entropy loss term $L_{M}$ measures the consistency between the mask and its groundtruth. Additionally, the loss term $L_{D}$ measures the consistency between the derived boundary of semantic mask and semantic boundary groundtruth. Consequently, the loss function $L_{m}$ to measure the total error of semantic segmentation task is updated as:
\begin{equation}
L_{m}=L_M+ \lambda_1 L_D
\label{loss_m}
\end{equation}
where $\lambda_1$ is a constant for balancing two losses.   \\
\noindent \textbf{Semantic Boundary Loss}  Compared with semantic segmentation, semantic boundary detection suffers more from the higher missing rate due to the sparsity of pixels around the boundary. In order to alleviate this impact,  we follow \cite{hed, casenet, steal} to  define the following class-balanced cross-entropy loss function $L_{2}$:
\begin{small}
	\begin{equation}
	L_E = -\sum_{k}^{K}\sum_{i}(\beta y_i^{k}logY_i^{k} + (1 - \beta) (1 - y_i^{k})log(1 - Y_i^{k}))
	\label{loss_2}
	\end{equation}
\end{small}
where $\beta$ is the percentage of non-edge pixels in the boundary groundtruth and $y_i^{k}$ is the groundtruth boundary label and binary indicating whether pixel $i$ belongs to class $k$. \\
\indent Thus, the integrated loss function $L_{total}$ is finally formulated as below:
\begin{equation}
L_{total}=L_{M} + \lambda_1 L_{D} + \lambda_{2} L_{E}
\label{loss_total}
\end{equation}
where $\lambda_2$ is a weight for balancing the boundary loss. In our experiments, we empirically set the two parameters $\lambda_1$ and $\lambda_2$ to 1 and 1000. 
\section{Experiments}
\subsection{Implementation details}
The proposed RPCNet is implemented with PyTorch. The channel number of feature map in the pyramid context module is set to 256. In the training step, we adopt data augmentation similar to \cite{ccnet,ocnet}. The base learning rate is set to 0.001 for all our experiments. Momentum and weight decay coefficients are set to 0.9 and 0.0001, respectively. We train our model with Synchronized BN \cite{inplaceabn} with $6\times$ NVIDIA 1080 Ti, and batch size is set to 6.  Random crops and horizontal flip is also applied. Training input resolution is set to $894 \times 894$. We train the dataset with 180 epochs. We optimize the network by using the ``poly'' learning rate policy where the initial learning rate is multiplied by $(1-\frac{iter}{max\_iter})^{power}$ with $power=0.9$. In the ablation experiments, we set the batch size to 4 and training epochs to 60 in order to speed up the training procedure. \\
\indent In the inference step, we follow \cite{encoding,danet,ccnet} to crop an image into several parts by using a sliding window to keep consistent with the training process.  Horizontal flipping is also adopted in the inference. We also use the multi-scale inference method with flipping when compared with state-of-the-art methods. In the ablation experiments, we use single-scale inference with flipping to evaluate.  For semantic boundary inference, we also follow \cite{steal} to use Test-NMS as a post-processing method to generate a more sharp boundary.
\begin{table}[t!]

	\centering
	\begin{tabular}{c|c|c|c|c}
		\toprule
		Duality loss& $\nabla M$ &PCM &mIoU  &MF (ODS) / AP \\ \hline\hline
		- & - & -                                              & 78.14& 73.61 /  72.81\\ 
		\checkmark & - & -                       &79.58& 74.24 / 73.57 \\ 
		\checkmark & \checkmark & - &  79.81&   74.45 / 74.20 \\ 
		\checkmark & \checkmark & $\{1\}$ & 79.92&  74.65 / 74.29  \\ 
		\checkmark & \checkmark & $\{1, 3\}$ & 80.20&  74.80 / 74.54\\ 
		\checkmark & \checkmark & $\{1, 3, 5, 7\}$ & 80.43& 75.54 / 75.14 \\ 
		\bottomrule
	\end{tabular}

	\caption{Ablation experiments for duality loss, $\nabla M$ fusion and pyramid context module (PCM). We set $S$ to 8 in the experiments.}
	\label{geo_eval}
\end{table}
\subsection{Dataset}
All of our experiments are conducted on the well-known Cityscapes dataset, which contains 2975 training, 500 validation and 1525 test images. Each image has a high resolution of  $2048 \times 1024$ pixels with 19 semantic classes. Noted that no coarse data is employed in our experiments.
We also follow \cite{casenet,steal} to generate the ground truth boundaries for semantic boundary detection task.
\begin{table}[t!]
	\centering
	\begin{tabular}{l|c|c}
		\toprule
		$S \quad$ & mIoU  &MF (ODS) / AP  \\ \hline\hline
		1 & 77.65 & - \\ 
		1 &  -    & 72.62 / 71.56 \\
		2 & 78.77 & 73.44 / 72.53 \\ 
		3 & 79.44 & 74.55 / 73.78 \\   
		4 & 79.80 & 74.56 / 73.80 \\ 
		5 & 79.88 & 74.61 / 73.91 \\ 
		6 & 80.25 & 74.94 / 74.31 \\ 
		7 & 80.36 &  75.10 / 74.38 \\ 
		8  &\textbf{80.43} &  \textbf{75.54} / \textbf{75.14} \\
		\bottomrule
	\end{tabular}
	\caption{Ablation experiments of iterative pyramid context module for semantic segmentation  and semantic boundary.  The iterative steps $S$ is set from 1 to 8. For $S=1$, only one task  is trained and evaluated.}
	\label{pcm_eval}
\end{table}
\subsection{Evaluation metric}
In this work, we use the mean intersection of union (mIoU) \cite{deeplab_v2,psanet,danet} to evaluate the semantic segmentation task, which denotes the ratio of correctly classified pixels in a class over the union set of pixels predicted to this class and groundtruth, and then averaged over all classes, i.e., $\frac{1}{N}\frac{t_{ii}}{T_{i}+\sum_{j}t_{ji}-t_{ii}}$. Here, $N$ is the number of semantic classes, and $T_i$ is the total number of pixels in class $i$, while ${t_{ij}}$ indicates the number of pixels which belong to class $i$ and predicted to class $j$. \\
\indent For semantic boundary detection, we follow the evaluation protocol proposed in \cite{seal,steal}, which is considerably harder than the one used in \cite{sem_boudary,casenet}. We report the maximum F-measure (MF) at optimal dataset scale (ODS), and average precision (AP) for each class. An essential parameter in the evaluation is the matching distance tolerance, which is defined as the maximum slack allowed for boundary predictions to be considered as correct matches to ground-truth. We follow \cite{steal} and set it to be 0.00375 in all our experiments. 
\begin{table}[t!]
	\centering
	\begin{tabular}{l|l|l}
		\toprule
		Method & Backbone &mIoU \\ \hline\hline
		DeeplabV2 \cite{deeplab_v2} & ResNet101 & 70.4\\ 
		Piecewise \cite{piecewise} & ResNet101 & 71.6 \\ 
		PSPNet \cite{PSPNet} & ResNet101 & 78.8 \\   
		DeeplabV3+ \cite{deeplabv3+} & ResNet101 & 78.8 \\ 
		InPlaceABN \cite{inplaceabn} & WideResNet38 & 79.4 \\ 
		GSCNN \cite{gate_scnn} & ResNet101 & 80.8 \\ 
		DANet \cite{danet} & ResNet101 & 81.5 \\ 
		\hline
	RPCNet (SS + Flip)   & ResNet101 &  81.8\\
	RPCNet (MS + Flip)   & ResNet101 &  \textbf{82.1} \\
		\bottomrule
	\end{tabular}
	
	\caption{Performance comparison between different strategies on Cityscape val set. ``SS'':  single scale test. ``MS'':  multi-scale test.}
	\label{cs_val}
\end{table}
\begin{table*}[t!]
	\resizebox{\textwidth}{!}{ %
		\centering
		\begin{tabular}{l|l|c|c|c|c|c|c|c|c|c|c|c|c|c|c|c|c|c|c|c|c}
			\toprule
			Method & Backbone  data &road &s.walk &build. &wall &fence &pole &t-light &t-sign  &veg &terrain &sky &person &rider &car &truck &bus &train &motor &bike &mean  \\ \hline\hline
			DeeplabV2 \cite{deeplab_v2} & ResNet101  &97.9& 81.3& 90.3& 48.8& 47.4 &49.6& 57.9& 67.3& 91.9 &69.4 &94.2& 79.8& 59.8&93.7& 56.5 		&67.5& 57.5& 57.7& 68.8 & 70.4\\
			RefineNet \cite{refinenet}& ResNet101  & 98.2& 83.3 &91.3& 47.8& 50.4 &56.1& 66.9& 71.3& 92.3& 70.3 &94.8 &80.9& 63.3& 94.5& 64.6& 76.1 &64.3 &62.2 &70.0 &73.6\\
			PSPNet \cite{PSPNet} & ResNet101 &98.6& 86.2& 92.9& 50.8& 58.8& 64.0& 75.6& 79.0& 93.4 &72.3& 95.4& 86.5& 71.3& 95.9& 68.2& 79.5& 73.8& 69.5 &77.2&78.4 \\
			AAF \cite{aaf} &ResNet101  & 98.5&  85.6 & 93.0&  53.8&  58.9 & 65.9 & 75.0&  78.4&  93.7 & 72.4 & 95.6&  86.4&  70.5&  95.9&  73.9&  82.7 & 76.9 & 68.7 & 76.4 &79.1\\
			DenseASPP \cite{denseaspp} &DenseNet161  & 98.7& 87.1& 93.4& 60.7 &62.7 & 65.6 & 74.6& 78.5& 93.6& 72.5& 95.4& 86.2& 71.9& 96.0& 78.0& 90.3& 80.7& 69.7 &76.8& 80.6\\
			PSANet \cite{psanet} & ResNet101 &- & - &- &- & - &- &-& -& -& -& - &- &-& - &- &-& -& -& -& 80.1\\
			SeENet \cite{SeENet} & ResNet101 & 98.7 & 87.3& 93.7 &57.1& 61.8& 70.5&77.6& 80.9& 94.0& 73.5& 95.9 &87.5& 71.6& 96.3& 76.4& 88.0& 79.9 &73.0& 78.5& 81.2\\
			ANNNet \cite{ANNNet} & ResNet101 &- & - &- &- & - &- &-& -& -& -& - &- &-& - &- &-& -& -& -& 81.3\\
			CCNet \cite{ccnet} & ResNet101 &- & - &- &- & - &- &-& -& -& -& - &- &-& - &- &-& -& -& -& 81.4\\
			BFP  \cite{BFP}&ResNet101 &  98.7& 87.0 &93.5 &59.8 &63.4& 68.9& 76.8& 80.9& 93.7& 72.8& 95.5& 87.0& 72.1& 96.0 &77.6& 89.0 &86.9& 69.2& 77.6 &81.4 \\
			DANet \cite{danet} &ResNet101 &98.6&  87.1 &93.5 &56.1 & 63.3 &69.7 &77.3& 81.3& 93.9& 72.9& 95.7 &87.3 &72.9& 96.2 &76.8 &89.4& 86.5& 72.2& 78.2& 81.5\\
			\hline 
			Ours &ResNet101  & 98.7 & 86.7 &93.9 & 62.4& 62.8 & 70.5 & 77.5 & 81.1 & 94.0 & 72.3 & 95.9 & 87.8 & 74.1 & 96.3 & 76.5 & 88.0 & 85.2 & 71.0 & 78.6 & \textbf{81.8}\\

			
			\bottomrule
		\end{tabular}
	}
	\caption{ Comparison vs state-of-the-art methods without coarse data training on the Cityscapes test set. }
\label{cs_test}
\end{table*}
\subsection{Ablation experiments}
Our proposed method models iterative pyramid context to interact between semantic segmentation and semantic boundary detection. For semantic segmentation, we propose duality loss to ensure boundary consistency between the predicted mask and groundtruth. For semantic boundary detection, we fuse the spatial gradient $\nabla M$ from semantic mask into the semantic boundary probability map to suppress non-edge pixels. In order to verify the effect of the proposed components in this paper, we perform detailed ablation experiments to compare the performance after using or removing these modules. We summarize the results in Table \ref{geo_eval} and Table \ref{pcm_eval}. \\
\noindent  \textbf{Duality Loss and $\nabla M$ Fusion} In Table \ref{geo_eval}, we first remove duality loss, spatial gradient $\nabla M$ fusion and pyramid context module to achieve mIoU score of 78.14\%   for semantic segmentation and MF / AP score of 73.61\% /  72.81\%  for semantic boundary detection while $S$ is set  to 8.   Duality  loss  brings 1.44\% improvement for semantic segmentation and 0.82\% / 0.76\% improvement for semantic boundary detection. The $\nabla M$ fusion also benefits the  two tasks, which  obtains mIoU score of 79.81\% for semantic segmentation and MF (ODS) / AP score of 74.45\% / 74.20\%  for semantic boundary detection.
\begin{figure}[t]
	\centering
	\includegraphics[width=0.96\linewidth]{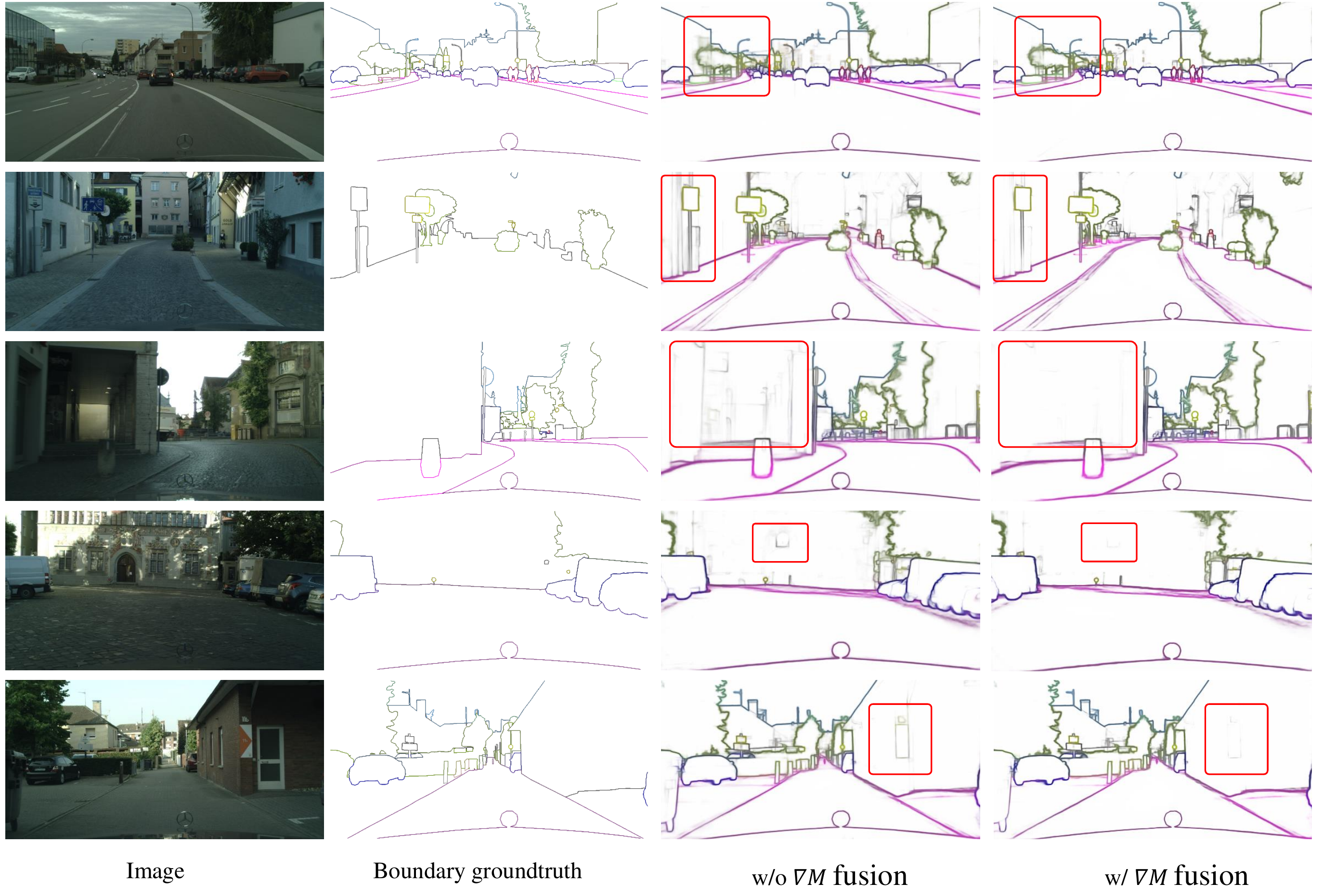}
	\caption{Some visualization comparison examples for  semantic boundary detection with or without $\nabla M$ fusion (\textbf{best viewed in color}). }
	\label{m_fuse}
\end{figure}
\begin{figure}[t!]
	\centering
	\includegraphics[width=0.96\linewidth]{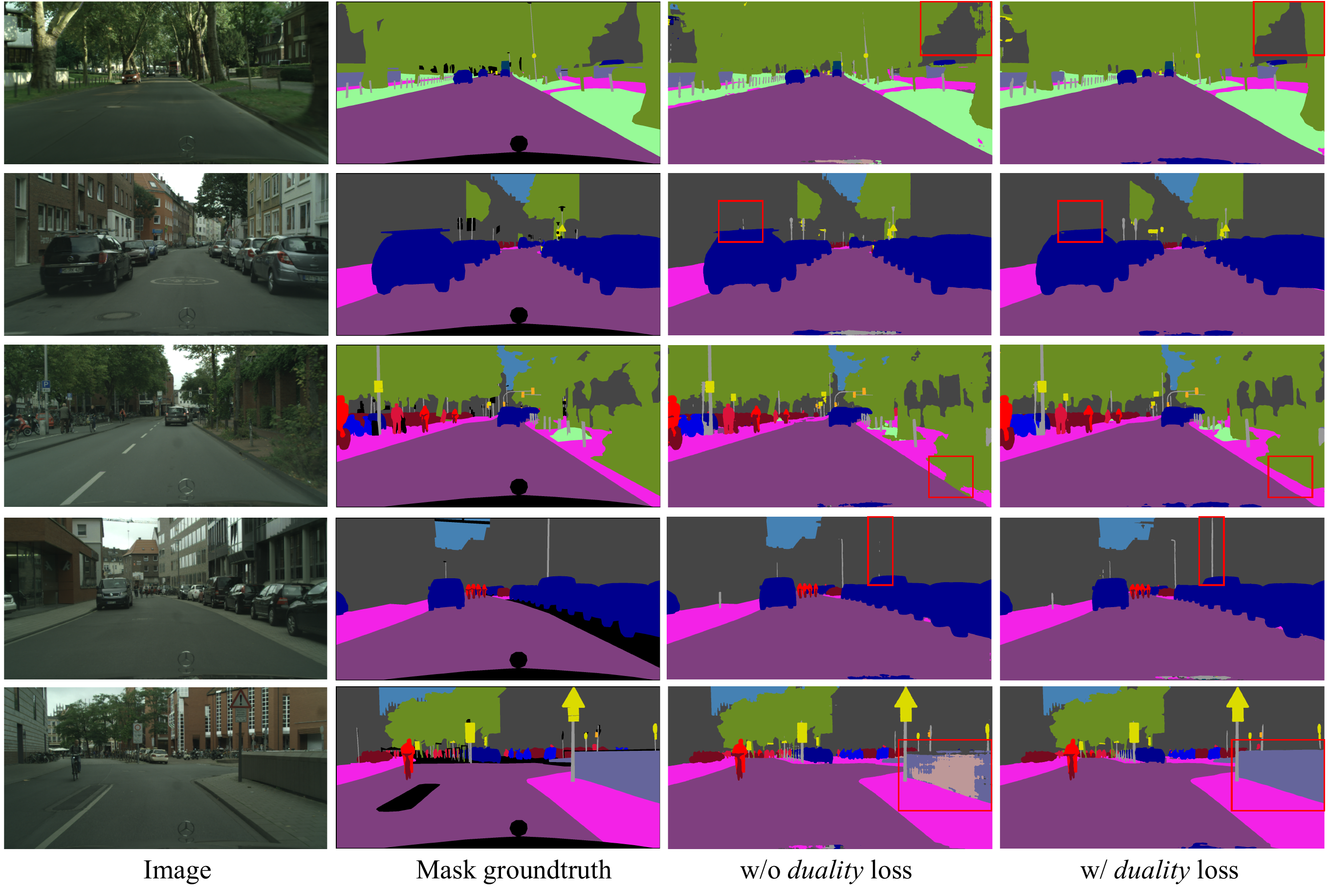}
	\caption{Some visualization comparison examples for  semantic segmentation with or without duality loss used (\textbf{best viewed in color}). }
	\label{geo_loss1}
\end{figure}\\
\noindent \textbf{Pyramid Context Module} For the pyramid context module, we first compare different patch partition methods as shown in Table \ref{geo_eval} and the patch partition setup ($1 \times 1$, $3 \times 3$, $5 \times 5$ and $7 \times 7$) achieves the best performance, which will be used in all other experiments. The pyramid context module boosts information exchange between semantic segmentation and semantic boundary detection. Compared with the setup without PCM, the PCM embedding can bring 0.62\% improvement for semantic segmentation, 1.09\% improvement on MF and 0.94\% improvement on AP  for semantic boundary detection. \\
\noindent \textbf{Iterative Pyramid Context Module} In Table \ref{pcm_eval},  we first report the performance for single task setup ($S=1$). It can be observed that combining two tasks together ($S=2$) benefits both the two tasks. We also present the results for semantic segmentation and semantic boundary detection on different $S$. As the iterative pyramid context module can help to refine the feature maps with each other, we can see that increasing step number $S$ can bring constant improvement for both tasks.  When $S=8$, we can achieve mIoU score of 80.43\% on semantic segmentation and MF  / AP of 75.54\% / 75.14\% on semantic boundary detection. Thus, we set $S$ to 8 in our subsequent experiments.
\\
\begin{figure*}[t!]
	\centering
	\includegraphics[width=0.94\linewidth]{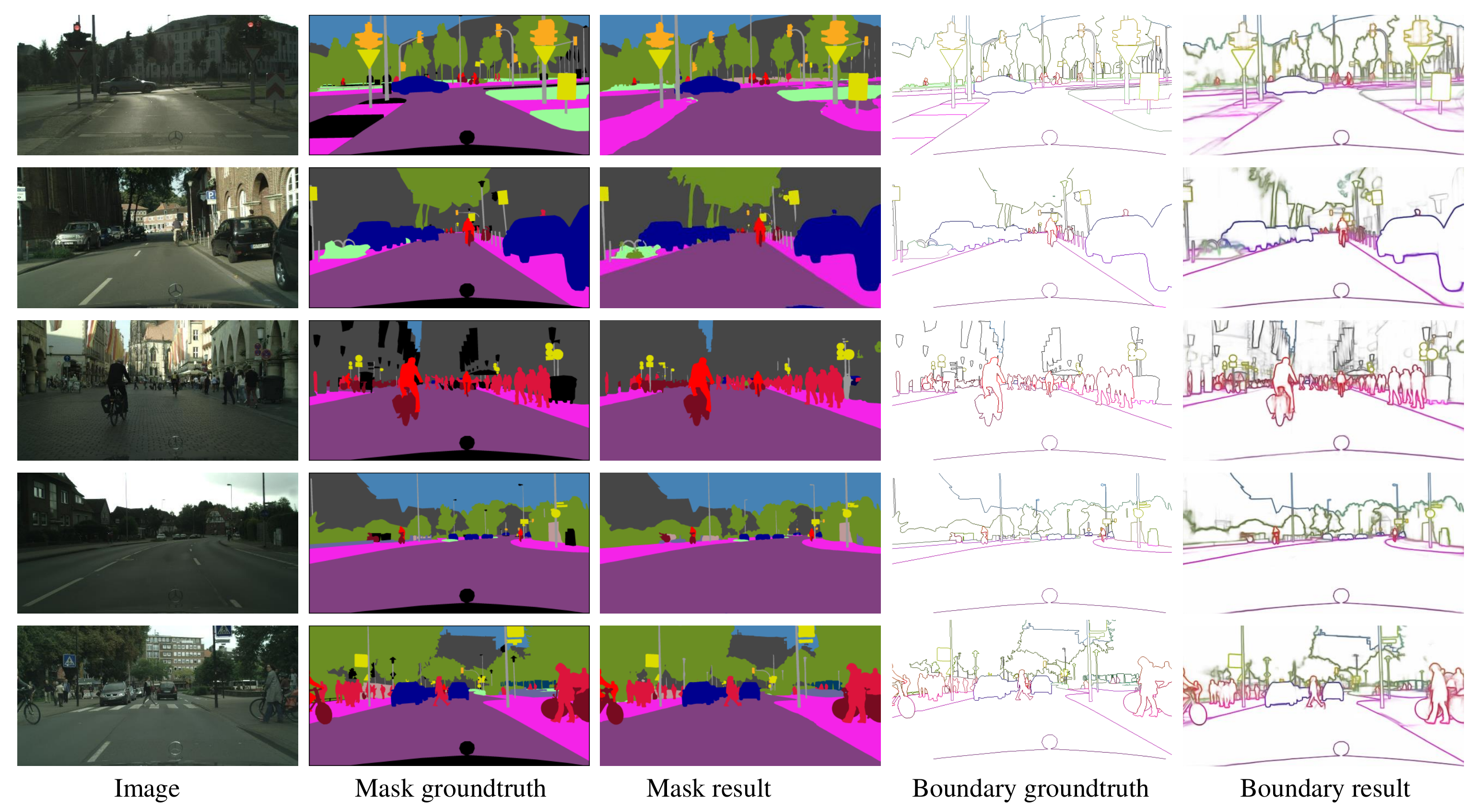}
	\caption{Visualization of results for semantic segmentation and semantic boundary detection (\textbf{best viewed in color}).}
	\label{vis_all}
\end{figure*}
\begin{table*}[t!]
	\resizebox{\textwidth}{!}{ %
		\centering
		\begin{tabular}{l|l|c|c|c|c|c|c|c|c|c|c|c|c|c|c|c|c|c|c|c|c|c}
			\toprule
			\textbf{Metric} &Method &Test NMS &road &s.walk &build. &wall &fence &pole &t-light &t-sign  &veg &terrain &sky &person &rider &car &truck &bus &train &motor &bike &mean  \\ \hline\hline
		
			\multirow{5}*{MF (ODS)}&CASENet \cite{casenet}&  &87.06 &75.95& 75.74& 46.87& 47.74& 73.23& 72.70& 75.65 &80.42& 57.77& 86.69 &81.02& 67.93 &89.10& 45.92& 68.05& 49.63& 54.21 &73.74& 68.92 \\
			
			~&$\text{CASENet}^{*}$ \cite{steal} &    &87.23 &76.08 &75.73 &47.86 &47.57& 73.67& 71.77 &75.19 &80.58& 58.39 &86.78& 81.00& 68.18& 89.31 &48.99 &67.82& 50.84& 55.30& 74.16& 69.29  \\
			
			~&$\text{CASENet}^{*}$  \cite{steal} & \checkmark   &88.13& 76.53 &76.75& 48.70& 48.60 &74.21& 74.54& 76.38 &81.32 &58.98& 87.26& 81.90 &69.05 &90.27& 50.93 &68.41 &52.11 &56.23 &75.66 &70.31 \\

			~&STEAL  \cite{steal}  &   &88.08& 77.62 &77.08& 50.02& 49.62& 75.48& 74.01& 76.66& 81.51& 59.41 &87.24& 81.90& 69.87& 89.50& 52.15& 67.80& 53.60 &55.93& 75.17 &70.67 \\
			~&STEAL  \cite{steal}  &\checkmark  &88.94 &78.21 &77.75& 50.59 &50.39& 75.54& 76.31& 77.45& 82.28 &60.19& 87.99 &82.48 &70.18 &90.40& 53.31& 68.50 &53.39 &56.99 &76.14& 71.42\\
			~&Ours    &\checkmark  &\textbf{90.86} & \textbf{82.32} &\textbf{ 82.11} & \textbf{57.15} & \textbf{58.97} &\textbf{84.48} & \textbf{83.34} & \textbf{82.26} & \textbf{84.88} & \textbf{64.22} & \textbf{89.87} & \textbf{86.28} & \textbf{78.47} & \textbf{92.61} & \textbf{67.75} & \textbf{82.79} & \textbf{68.48} & \textbf{69.20} & \textbf{80.09} & \textbf{78.22} \\
			\hline\hline
			
			\multirow{5}*{AP}&CASENet \cite{casenet}     &    &54.58 & 65.44  &67.75  &37.97  &39.93 & 57.28  &64.65  &69.38 & 71.27 & 50.28  &73.99 & 72.56  &59.92  &66.84  &35.91 & 56.04  &41.19  &46.88  &63.54 & 57.65 \\
			~&$\text{CASENet}^{*}$ \cite{steal}      &      &68.38 &69.61 &70.28& 40.00& 39.26& 61.74& 62.74 &73.02 &72.77 &50.91& 80.72& 76.06& 60.49 &79.43 &40.86& 62.27& 42.87& 48.84 &64.42 &61.30\\
			
			~&$\text{CASENet}^{*}$ \cite{steal}      &   \checkmark   &88.83 &73.94 &76.86& 42.06 &41.75& 69.81& 74.50& 76.98 &79.67 &56.48& 87.73& 83.21& 68.10 &91.20& 44.17 &66.69 &44.77& 52.04& 75.65& 68.13\\
	
			~&STEAL  \cite{steal}  &   &89.54& 75.72& 74.95 &42.72 &41.53& 65.86& 67.55 &75.84 &77.85& 52.72& 82.70& 79.89& 62.59& 91.07 &45.26& 67.73 &47.08& 50.91& 70.78& 66.44 \\
			~&STEAL  \cite{steal}  & \checkmark  &90.86 &78.94& 77.36& 43.01& 42.33 &71.13& 75.57 &77.60 &81.60 &56.98 &87.30& 83.21 &66.79 &91.59& 45.33& 66.64& 46.25 &52.07& 74.41& 68.89  \\	
			~& Ours & \checkmark & \textbf{91.27} & \textbf{83.87} & \textbf{84.00} &\textbf{ 53.18} & \textbf{54.96} &\textbf{ 84.55} &\textbf{ 85.48} & \textbf{84.66} &\textbf{ 86.15} & \textbf{61.18} &\textbf{ 90.7}2 &\textbf{ 88.95} &  \textbf{79.95} & \textbf{94.40} & \textbf{68.11} & \textbf{85.47} & \textbf{68.53} & \textbf{69.44} & \textbf{82.17} & \textbf{78.79}\\
			\bottomrule
		\end{tabular}
	}	
	\caption{Quantitative results on the val set on the Cityscapes dataset. We use ResNet101 pretrained on ImageNet as backbone. $\text{CASENet}^{*}$  is the reimplementation of CASENet in \cite{steal}.  Scores are measured by $\%$.}
	\label{cs_boundary}
\end{table*}
\subsection{Comparison with state-of-the-art works}
\noindent \textbf{Semantic Segmentation} We first summarize state-of-the-art results on Cityscapes validation set in Table \ref{cs_val}. We achieve the best performance compared with these methods. In particular, we outperform GSCNN by 1.3\%, which leverages the binary edge as gate to boost the performance. \\ 
\indent We also compare our method with state-of-the-art methods on Cityscapes test set. Specifically, we finetune our best model RPCNet with only fine annotated trainval data, and submit our test results to the official evaluation server.  Results are shown in Table \ref{cs_test}. We can see that our RPCNet achieves a new state-of-the-art performance of 81.8\% on the test set. With the same backbone ResNet-101, our model outperforms DANet\cite{danet}. Besides, RPCNet also surpasses BFP \cite{BFP}, which makes use of binary edge information to propagate local features within their regions. \\
\noindent \textbf{Semantic Boundary Detection} We compare  RPCNet with state-of-the-art methods for semantic boundary detection on Cityscapes validation set in Table \ref{cs_boundary}.  STEAL \cite{steal}  proposes a novel loss to enforce the edge detector to predict a maximum response along the normal direction at edges,  which is the current state-of-the-art method. However, we propose a better method to suppress non-semantic edges and attain high-quality representation. Our method achieves new state-of-the-art results over previous works by a large margin on both MF (ODS)  metric and AP metric.
\subsection{Visualization Results}
To better understand the effect of the proposed methods, we present some visual examples. As shown in Figure \ref{m_fuse}, the visual examples for semantic boundary detection with or without $\nabla M$ are compared. We can see that after $\nabla M$ fusion, we can obtain more accurate semantic boundaries. The pixels, even though they belong to edges but not semantic boundaries, are suppressed so that we can locate pixels belonging to semantic boundaries. We also visualize the examples for semantic segmentation with or without duality loss in Figure \ref{geo_loss1}. It can be observed that the novel loss can help to generate more precise boundaries. The  misclassified ``Pole''  is also recognized after duality loss is used. At last, we also visualize some examples for both semantic segmentation and semantic boundary detection.  In Figure \ref{vis_all}, the mask outputs are accurate  and the quality of the semantic boundaries is also very high.
\section{Conclusion}
In this paper, we have presented a joint-task framework for both semantic segmentation and semantic boundary detection. We use an iterative pyramid context from one task at multiple scales to refine the feature map of another task alternately, which helps the two tasks interact with each other. In order to resolve the sparse boundary issues, we fuse derived boundary from semantic segmentation mask into semantic boundary probability map to suppress non-semantic edge pixels.  The novel loss function originated from the dual constraint  is designed to improve further the performance for semantic segmentation, which ensures the consistency between semantic mask boundary and boundary groundtruth. The comprehensive experiments on the Cityscapes dataset verify the effectiveness of the proposed  framework and show that proposed RPCNet outperforms current state-of-the-art works not only on semantic segmentation task but also on semantic boundary detection task.
\section{Acknowledgments}
This work is supported by  Hong Kong RGC GRF 16206819, Hong Kong RGC GRF 16203518, Hong Kong T22-603/15N.
{\small
	\bibliographystyle{unsrt}
\bibliography{banet}
}

\end{document}